%% file: main.tex
\documentclass[10pt,twocolumn,letterpaper]{article}

\usepackage{cvpr}              
\input{preamble}
\definecolor{cvprblue}{rgb}{0.21,0.49,0.74}
\usepackage[pagebackref,breaklinks,colorlinks,allcolors=cvprblue]{hyperref}

\usepackage{graphicx}
\usepackage{float}
\usepackage[utf8]{inputenc} 
\usepackage[T1]{fontenc}    
\usepackage{url}            
\usepackage{booktabs}       
\usepackage{amsfonts}       
\usepackage{nicefrac}       
\usepackage{microtype}      
\usepackage{xcolor}         
\usepackage{multirow}
\usepackage{makecell}
\usepackage{caption}

\usepackage{amsmath}
\usepackage{amssymb}
\usepackage{colortbl}
\usepackage{array} 

\usepackage{newfloat}
\usepackage{listings}

\usepackage[ruled,vlined]{algorithm2e}
\usepackage{algpseudocode}

\def\paperID{} 
\def\confName{CVPR}
\def\confYear{2026}

\title{Hypergraph-State Collaborative Reasoning for Multi-Object Tracking}

\author{
Zikai Song\textsuperscript{\rm 1, 2}, 
Junqing Yu\textsuperscript{\rm 1}\footnotemark[1] ,
Yi-Ping Phoebe Chen\textsuperscript{\rm 3}, 
Wei Yang\textsuperscript{\rm 1},
Xinchao Wang\textsuperscript{\rm 2}\footnotemark[1]\\[1ex]
\textsuperscript{\rm 1}Huazhong University of Science and Technology\qquad
\textsuperscript{\rm 2}National University of Singapore \\
\textsuperscript{\rm 3}La Trobe Uniersity
}

\begin{document}
\maketitle
\makeatletter
\def\@makefntext#1{\noindent #1}  
\footnotetext{* Corresponding Authors: Xinchao Wang (xinchao@nus.edu.sg) and 
Junqing Yu (yjqing@hust.edu.cn)}
\footnotetext{Code is available at \underline{https://github.com/SkyeSong38/HyperMOT}}
\makeatother

\input{sec/0_abstract}    
\input{sec/1_intro}

\input{sec/2_formatting}
\input{sec/3_finalcopy}

{
    \small
    \bibliographystyle{ieeenat_fullname}
    \bibliography{main}
}



\end{document}

%% file: sec/0_abstract.tex
\begin{abstract}

Motion reasoning serves as the cornerstone of multi-object tracking (MOT), as it enables consistent association of targets across frames. However, existing motion estimation approaches face two major limitations: (1) instability caused by noisy or probabilistic predictions, and (2) vulnerability under occlusion, where trajectories often fragment once visual cues disappear.
To overcome these issues, we propose a \textbf{collaborative reasoning} framework that enhances motion estimation through joint inference among multiple correlated objects. By allowing objects with similar motion states to mutually constrain and refine each other, our framework stabilizes noisy trajectories and infers plausible motion continuity even when target is occluded.
To realize this concept, we design \textbf{HyperSSM}, an architecture that integrates Hypergraph computation and a State Space Model (SSM) for unified spatial–temporal reasoning. The Hypergraph module captures spatial motion correlations through dynamic hyperedges, while the SSM enforces temporal smoothness via structured state transitions. This synergistic design enables simultaneous optimization of spatial consensus and temporal coherence, resulting in robust and stable motion estimation.
Extensive experiments on four mainstream and diverse benchmarks(MOT17, MOT20, DanceTrack, and SportsMOT) covering various motion patterns and scene complexities, demonstrate that our approach achieves state-of-the-art performance across a wide range of tracking scenarios. 
\end{abstract}

%% file: sec/1_intro.tex
\section{Introduction}
\label{sec:intro}

\begin{figure}[!h]
    \centering
    \includegraphics[width=1\linewidth,keepaspectratio,page=1]{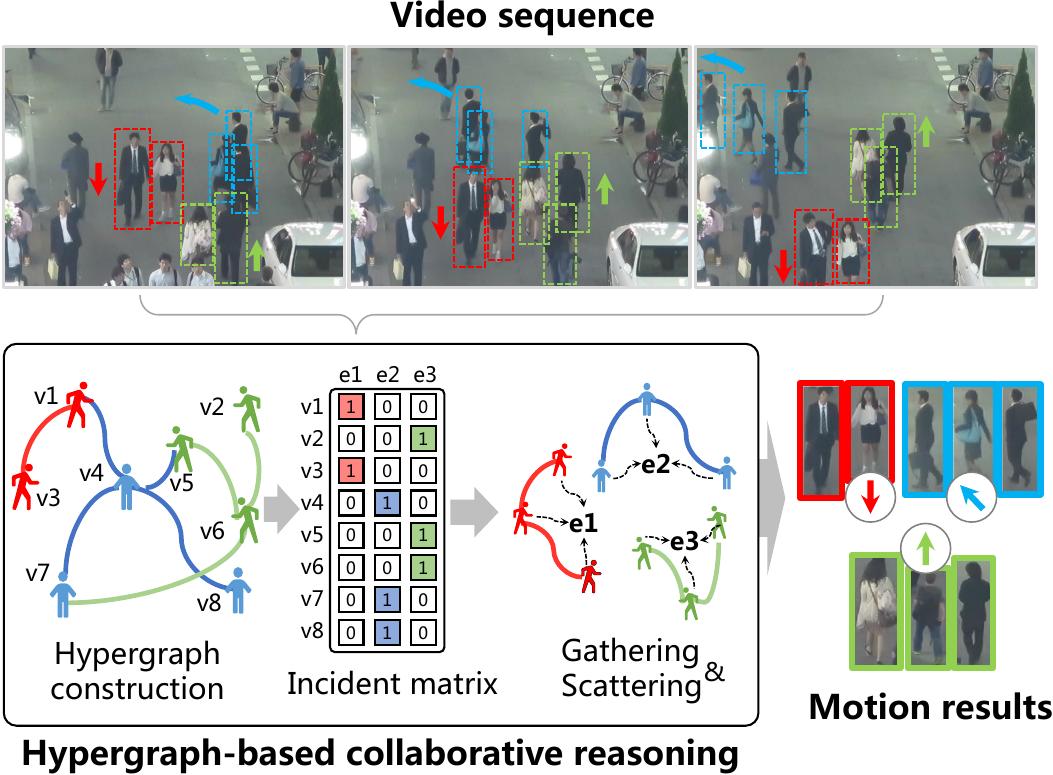}
    \caption{\textbf{Hypergraph-based collaborative motion estimation}. We first construct a motion‐aware hypergraph, represented by an incidence matrix, the hyperedge (\textbf{e}) selectively associating strongly correlated targets based on motion states (\textbf{gathering}), the gathered group is then dispersed back to individual nodes for fine-grained motion refinement (\textbf{scattering}), achieving both collaborative inference and precise object-specific estimation.}
  \label{fig:1}
\end{figure}

Multi-object tracking (MOT) aims to detect and continuously track multiple targets in dynamic scenes while preserving their identities across consecutive frames~\cite{TumTrack, Odtrack, mmstTrack, cadtrack, CSWinTT, compact, wang2015tracking, Ren2020}. Unlike object detection, which focuses solely on identifying objects within a single image~\cite{deep, improving, knowledge, counterfactual}, MOT must establish temporal associations between detections over time. This makes accurate estimation of inter-frame motion indispensable. Motion estimation thus lies at the core of MOT, serving as the fundamental mechanism that enables consistent target association and identity maintenance throughout a video sequence~\cite{adapTrack, unifiedTrack,DecoupledTrack, altrack}.

Existing MOT methods can be divided into two distinct categories depending on whether they explicitly estimate motion displacement. (1) \textbf{Query-based methods}~\cite{MOTR,MOTRv2,motip} eliminate the need for explicit motion modeling. They represent each target as a learnable query with each query corresponding to a specific object, through iterative network updates, the query directly regresses the object’s position. (2) \textbf{Motion-based methods} explicitly estimate object motion and integrate it with detection results for tracking. Traditional approaches~\cite{ByteTrack, sparsetrack} often employ \textit{\textbf{Kalman Filter motion estimation}}. Recent frameworks~\cite{trackssm, OFTrack}, further leverage \textit{\textbf{learnable motion estimation}} to capture complex, nonlinear motion dynamics, leading to more robust tracking performance.

Our approach follows the motion-based paradigm and aims to tackle two long-standing challenges in motion estimation. \textbf{(C1)} \textbf{\textit{Motion noise and instability}}. Paradoxically, despite the rapid development of learning-based approaches, the hand-crafted Kalman Filter (KF) and its variants remain among the most stable motion estimators. KF assumes constant-velocity linear motion and performs remarkably well in simple, linear trajectories (such as MOT17~\cite{MOT16}). In contrast, learnable motion estimators can model complex nonlinear dynamics, yet their inherently probabilistic predictions often introduce noise and fluctuations, leading to less consistent performance in linear scenarios. \textbf{(C2)} \textit{\textbf{Robustness under occlusion}}. When a target becomes heavily occluded, motion estimation tends to fail, resulting in trajectory fragmentation or identity switches. Both KF and learnable motion estimation methods struggle to maintain temporal continuity once visual evidence is lost.

To address these challenges, we introduce a collaborative reasoning framework for motion estimation, as shown in Figure~\ref{fig:1}. Unlike other approaches that estimate motion independently for each target, our method performs joint estimation across multiple correlated objects. Rather than aggregating all targets indiscriminately, we selectively associate those with highly correlated motion states (in terms of direction and velocity), to preserve individual motion characteristics while enabling cooperative inference.
Our approach builds on two key principles to tackle the aforementioned challenges.
(1) \textit{\textbf{Joint reasoning inherently stabilizes motion estimation}} (for \textbf{C1}). When the trajectory of one target becomes noisy or unstable, the shared constraints from other motion-correlated targets help suppress instability and restore consistency.
(2) \textit{\textbf{Cross-object motion transfer enhances occlusion resilience}} (for \textbf{C2}). When a target becomes temporarily invisible, its motion can be probabilistically inferred from the observable dynamics of other targets exhibiting similar motion patterns. (e.g. estimating the trajectory of an occluded pedestrian based on the motion trends of others moving in the same direction.)



We propose \textbf{HyperSSM} to realize collaborative reasoning for motion estimation, HyperSSM is a unified architecture that synergizes Hypergraph computation with State Space Model (SSM) for \textit{\textbf{spatial collaborative reasoning}} and \textit{\textbf{temporal motion estimation}}. 
(a) From the spatial perspective, the Hypergraph module dynamically constructs motion-aware hyperedges that group objects exhibiting correlated motion states. These hyperedges aggregate object-specific motion information into shared latent group states, enabling spatial collaboration that suppresses individual estimation noise through consensus reasoning.
(b) From the temporal perspective, the SSM module models motion evolution via a structured autoregressive process, where motion states are propagated through learnable transition matrices to maintain temporal continuity and adapt to dynamic changes.
(c) Finally, HyperSSM bridges spatial and temporal reasoning by embedding hypergraph-derived relational constraints directly into the SSM’s state transition rules. This integration enforces consistent optimization of spatial group coherence and temporal trajectory smoothness, resulting in robust and stable motion estimation.
Comprehensive experiments on both linear motion datasets (MOT17~\cite{MOT16}, MOT20~\cite{MOT20}) and non-linear motion datasets (DanceTrack~\cite{DanceTrack}, SportsMOT~\cite{sportsmot}) demonstrate the state-of-the-art performance of our method.

In summary, our main contributions are as follows:
\begin{enumerate}
\item We propose a collaborative motion estimation framework that performs joint reasoning among correlated objects, effectively suppressing noise fluctuations and enhancing robustness under occlusion through cross-object motion constraint and information sharing.
\item We design HyperSSM, a unified architecture that integrates Hypergraph computation with SSM to achieve spatial–temporal collaborative reasoning. The Hypergraph module dynamically captures motion correlations among objects, while the SSM propagates these relationships through structured temporal transitions, jointly optimizing spatial coherence and temporal smoothness for stable motion estimation.
\end{enumerate}

%% file: sec/2_formatting.tex
\section{Related Work}

Object tracking has numerous applications in computer vision~\cite{MIPR, SSET, coupledmamba, mdvlad, mvp, finegrain, distractor}. Existing MOT methods can be divided into Query-based methods and Motion-based methods depending on whether they explicitly estimate motion of objects.

\vspace{5pt} \noindent \textbf{Query-based methods} (such as TrackFormer~\cite{TrackFormer} and the MOTR series~\cite{MOTR, MOTRv2, motrv3}) do not require explicit motion computation. Derived from DETR~\cite{DETR, DeformableDETR}, they represent each target as a learnable query that regresses object positions, maintaining identity consistency through persistent queries across the sequence. Recently, when enhanced with historical memory and large-scale hybrid modeling~\cite{motrv3,samba,sam2mot,motip}, these methods have achieved strong performance in highly nonlinear and number-fixed scenarios such as DanceTrack~\cite{DanceTrack}.

However, query-based methods have inherent limitations. Because each query is fixed to a specific target, they struggle to adapt to object appearances and disappearances, performing well mainly in scenarios with a constant number of targets. In dynamic scenes, their tracking stability degrades, and the query-to-box regression often introduces noisy fluctuations, leading to inferior performance in linear-motion benchmarks such as MOT17~\cite{MOT16} and MOT20~\cite{MOT20}.

\vspace{5pt} \noindent \textbf{Motion-based methods} decouple detection and association. They first predict object bounding boxes in each frame using object detectors~\cite{FasterRCNN, YOLOX}, then explicitly estimate object motions, and finally associate each detection with its most likely trajectory. (1) Many trackers~\cite{SORT, DeepSORT, ByteTrack, BoTSORT, hybridSORT} employ hand-crafted \textbf{\textit{Kalman Filters for motion estimation}}, which assume constant-velocity linear motion. Such methods perform remarkably well in linear-motion scenarios like surveillance videos but deteriorate sharply in nonlinear cases such as dancing or sports. (2) Recently, several approaches~\cite{trackssm, mambatrack, trackmamba, mambatrackpp, OFTrack} have introduced \textit{\textbf{learnable motion estimation networks}}, replacing the Kalman filter with data-driven predictors that better capture complex nonlinear dynamics and substantially improve tracking robustness.

Our method follows this learnable motion estimation paradigm, which combines the strengths of query-based and Kalman-based approaches. By decoupling detection and motion estimation, it avoids the fixed-number and linear-motion limitations of query-based methods, while the learnable motion predictor overcomes the Kalman filter’s weakness in nonlinear dynamics. However, current learnable estimators still suffer from noisy predictions and poor robustness under occlusion. To address these issues, we introduce a collaborative reasoning framework that jointly estimates motion among correlated targets, effectively stabilizing trajectories and enhancing occlusion resilience.

\vspace{5pt} \noindent \textbf{Graph learning in MOT.}
In recent years, several methods~\cite{star, SUSHI, TrackMPNN, GMTracker, graphmfn} have introduced graph neural networks (GNNs)~\cite{gnn1} to perform association matching in multi-object tracking. These methods model objects across frames as nodes, while edges represent potential associations between detections. By learning relational dependencies among nodes, GNNs infer the likelihood that two detections belong to the same target, thereby enabling accurate cross-frame matching and identity association.

Existing graph-based MOT methods mainly focus on linking the same target across consecutive frames to achieve identity association, but they still track each object separately without considering relationships among different targets.
In contrast, our approach models motion correlations among multiple objects within the same frame through a hypergraph structure, enabling collaborative tracking by capturing complex, high-order interactions.
This design introduces a new paradigm for MOT, allowing the model to reason collectively over multiple targets rather than treating them as independent entities.

\begin{figure*}[!h]
    \centering
    \includegraphics[width=0.9\linewidth,keepaspectratio,page=2]{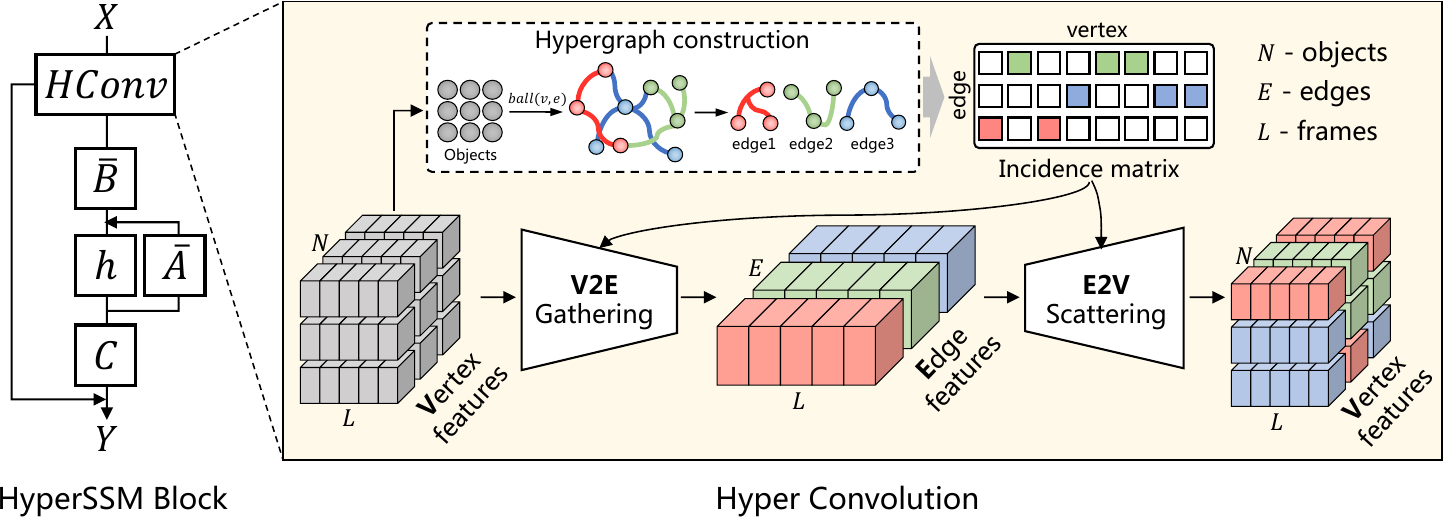}
    \caption{\textbf{HyperSSM block and Hyper Convolution.} The HyperSSM block integrates hypergraph computation into the SSM. Given the input motion feature $X$ of all objects $N$ within $L$ frames. First, collaborative reasoning is performed via hyper convolution (HConv), involving a \textit{vertex-to-edge (V2E) gathering} from vertices to hyperedges and followed by a \textit{edge-to-vertex (E2V) scattering} back to vertices. The block then computes hidden states $h$ from the HConv-processed input $X$ using discretized parameters $\overline{A}$ and $\overline{B}$. Finally, outputs $Y$ are generated through trainable parameters $C$ and an additional HConv block.}
  \label{fig:2}
\end{figure*}

\section{Method}



In this section, we present our tracking framework following a bottom-up exposition. 

\begin{itemize}
\item We first introduce the \textbf{core component HyperSSM block} (Sec. \ref{HyperSSM}), which details how multiple targets are aggregated and jointly reasoned to enable collaborative modeling. To clarify its design, we begin with the necessary preliminaries (Sec. \ref{Preliminary}) covering the concepts of Hypergraph and State Space Model (SSM).

\item We then describe how HyperSSM blocks are integrated into a \textbf{motion estimation module} (Sec. \ref{motion}) to perform motion prediction.

\item Finally, we outline the \textbf{overall tracking framework} (Sec. \ref{overall}), which integrates all components into a complete motion-based tracking pipeline.
\end{itemize}


\subsection{Preliminary}
\label{Preliminary}

\vspace{5pt} \noindent \textbf{Hypergraph Learning.} 
Unlike conventional graphs where each edge connects only two nodes, a \textbf{hypergraph} allows a single hyperedge to connect multiple nodes, thereby capturing complex and high-order relationships. 
Hypergraph computation~\cite{18,28,hyperyolo} generally consists of two stages: 
(1) \textbf{Construction.} Given a feature map $X$ extracted from a neural network, the construction stage applies a mapping function $f : X \rightarrow G$ to build a hypergraph $G$ by inferring latent correlations among features. 
(2) \textbf{Convolution.} The subsequent hypergraph convolution propagates high-order information through \textit{\textbf{gathering}} and \textit{\textbf{scattering}} operations defined as:
\begin{equation}
\label{eq:hyper}
\begin{aligned}
    Y &\xleftarrow{\text{gathering}} WD_e^{-1}H^{\!T}X, \\
    X_h &\xleftarrow{\text{scattering}} \sigma(D_v^{-1}HY\Theta),
\end{aligned}
\end{equation}
where $Y$ denotes the hyperedge features, and $W$ is a diagonal matrix of edge weights. 
$D_e$ and $D_v$ represent diagonal matrices of edge and vertex degrees, respectively, while $H$ is the $|v|\times|e|$ incidence matrix of the hypergraph. 
$\Theta$ denotes learnable parameters, and $\sigma(\cdot)$ is a nonlinear activation function.

\vspace{5pt} \noindent \textbf{State Space Model.} 
State Space Models (SSMs)~\cite{mamba, mamba2} have demonstrated strong capability in modeling long-range dependencies within sequential data. 
They are derived from the concept of continuous dynamical systems by introducing a hidden state $h_t$ that maps a sequence of inputs $x_t$ to corresponding outputs $y_t$, where $N$ denotes the dimension of the hidden state. 
Among recent variants, the S4 model~\cite{s4, s4d} is one of the most widely used formulations, whose autoregressive process can be expressed as:
\begin{equation}
\label{eq3}
\begin{aligned}
    h_{t} &= \mathbf{\overline{A}}h_{t-1}+\mathbf{\overline{B}}x_{t},\\
    y_{t} &= \mathbf{C}h_{t} + \mathbf{D}x_t,
\end{aligned}
\end{equation}
where $x_t$ denotes the input, $h_t$ is the hidden state, and $y_t$ is the output. 
$\mathbf{\overline{A}}$, $\mathbf{\overline{B}}$, $\mathbf{C}$, and $\mathbf{D}$ are learnable matrices, with $\mathbf{\overline{A}}$ and $\mathbf{\overline{B}}$ typically discretized from continuous parameters using zero-order hold (ZOH) or Euler discretization rules.


\subsection{HyperSSM block}
\label{HyperSSM}

\vspace{5pt} \noindent \textbf{Hypergraph Construction}. We define a hypergraph $G = \{V , E \}$ by its vertex set $V$ and hyperedge set $E$. In our approach, we define each object is represented as a vertex, and its motion feature $x\in\mathbb{R}^d$ constitutes the vertex set $V$. To model the relationships among objects with similar motion status, a distance threshold $\theta$ is used to construct an $e$-ball around each motion feature, which will serve as a hyperedge. The overall hyperedge set is defined as $E=\{ball(v,e)~|~v\in V\}$, where $ball(v,e)=\{u| ~||x_u-x_v||_d<\theta,~u\in V\}$ denotes the set of neighboring vertex for a given $v$. In computations, a hypergraph $G$ is typically represented by its incidence matrix $H$.

Computing \textbf{motion feature} $x \in \mathbb{R}^d$ is as follows: 
Let the historical trajectory of an object be denoted as $\mathbf{p}_L \in \mathbb{R}^{L \times 4}$, where $L$ is the number of frames and each element $p_t = (x, y, w, h)$ represents the object’s box at frame $t$. 
For each target, we first compute the displacement between consecutive frames to obtain the velocity sequence:  
\[
v_t = p_t - p_{t-1}, \quad t = 2, \ldots, L.
\]
To emphasize recent motion trends while suppressing noise, we apply an \textit{exponential moving average (EMA)} weighting scheme, where frames closer to the current one receive higher weights:  
\[
\hat{v} = \sum_{t=1}^{L} \alpha (1-\alpha)^{L-t} v_t,
\]
with $\alpha$ controlling the decay rate. 
Finally, the smoothed velocity $\hat{v}$ is passed through a \textit{trajectory embedding module} to obtain the motion representation $x \in \mathbb{R}^d$.

\vspace{5pt} \noindent \textbf{Hyper Convolution (\textit{HConv}).} We employ a standard spatial-domain Hyper convolution \cite{18} with an additional residual connection. This process facilitates collaborative reasoning through a gathering phase from vertices to hyperedges and a scattering phase from hyperedges to vertices, as described below:
\begin{equation}
\label{eq:hyperconv1}
\left\{
\begin{aligned}
x_e &= \frac{1}{|N_v(e)|}\sum_{v\in N_v(e)}x_v\Theta \\
x_v' &= x_v+\frac{1}{|N_e(v)|}\sum_{e\in N_e(v)}x_e
\end{aligned}
\right.
\end{equation}
where $N_v(e)$ and $N_e(v)$ are two indicate functions defined as: $N_v(e)=\{v| v\in e, v\in V\}$ and $N_e(v)=\{e|v\in e,e\in E\}$. $\Theta$ is a trainable parameter. The matrix formulation of the hypergraph computation can be defined as:

\begin{equation}
\label{eq:hyperconv2}
\begin{aligned}
HConv(X) &= \sigma(D^{-1}_vHWD^{-1}_eH^TX_v\Theta)
\end{aligned}
\end{equation}
where $D_v$ and $D_e$ represent the diagonal degree matrices of the vertices and hyperedges.

\vspace{5pt} \noindent \textbf{\textit{HConv}-embedded SSM.}
Our HyperSSM block integrates Hypergraph structure into State Space Model, as illustrated in Figure~\ref{fig:2}. Building upon the conventional SSM workflow where the \textit{\textbf{first step}} computes hidden states $h_t$ from input $X_t$ in frame $t$ using discretized parameters $\overline{A}$ and $\overline{B}$, and the \textit{\textbf{second step}} generates outputs $Y_t$ via trainable parameters $C$ and $D$. HyperSSM introduces three critical modifications: (1) embedding Hypergraph computation during the initial state transition phase to model multi-object motion interactions within $X_t$, (2) removing the redundant $D$ parameter to prevent over-parameterization while leveraging trainable hypergraph parameters $\Theta$ from the hypergraph computation, and (3) incorporating a residual connection that combines the original input $X_t$ with the transformed hidden state to preserve essential motion features. The operation can be formalized as:
\begin{equation}
\label{eq:hyssm}
\begin{aligned}
    h_{t} &= \mathbf{\overline{A}}h_{t-1}+\mathbf{\overline{B}}\cdot HConv(X_{t})\\
    Y_{t} &= X_{t} + \mathbf{C}h_{t} + HConv(X_{t}).
\end{aligned}
\end{equation}

These innovations enhance spatiotemporal relationship modeling through Hypergraph-structured reasoning while retaining the computational efficiency of SSM, enabling robust motion estimation with explicit capture of complex object interactions and streamlined parameterization.

\begin{figure}[!h]
    \centering
    \includegraphics[width=1\linewidth,keepaspectratio,page=3]{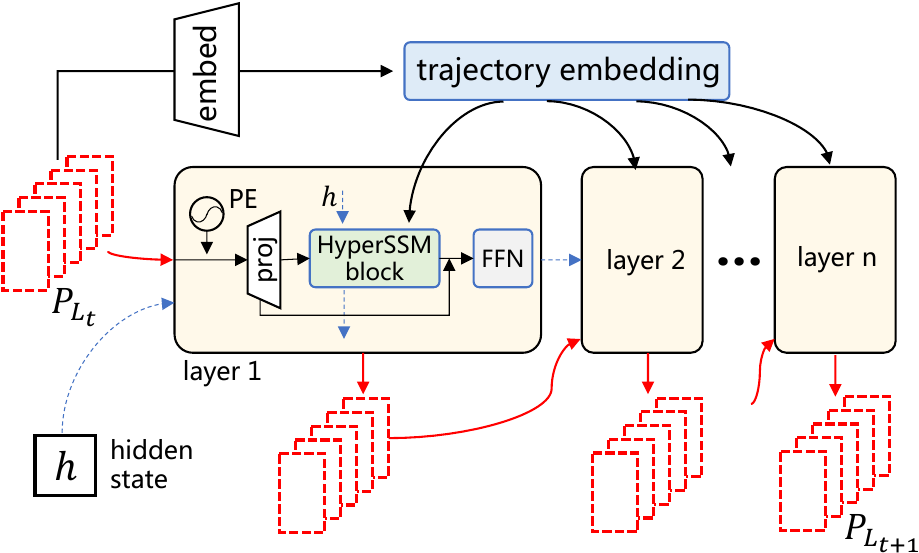}
    \caption{\textbf{The architecture of the motion estimation model.} The motion estimation module comprises multiple cascaded HyperSSM layers. Given the multi-frame multi-object position information $\mathbf{P}_{L_{t}}$, we encode it into trajectory embeddings to guide each layer. The resulting motion features are processed through a feed-forward network (FFN), generating the position information $\mathbf{P}_{L_{t+1}}$ for the next sliding window at one-frame intervals.}
  \label{fig:3}
\end{figure}

\subsection{Motion Estimation}
\label{motion}
The motion estimation module is illustrated in Figure~\ref{fig:3}. Given the position information $\{P^i_t=(x,y,w,h)\}^{i\in N}_{t \in L}$ of all objects $N$ across $L$ frames in the range $[t-L+1, t]$, we encode this multi-frame multi-object position information $\mathbf{P}_L^N \in \mathbb{R}^{N\times L \times 4}$ into trajectory embeddings $\mathcal{T}_L^N \in \mathbb{R}^{N\times L \times d}$. These trajectory embeddings $\mathcal{T}$ are used as guidance and fed, along with the position information, into a motion estimation module composed of multiple cascaded layers, each layer containing a HyperSSM block. The resulting motion features are passed through a feed-forward network (FFN) to estimate the motion result $\mathbf{P}_{L_{t+1}}^N $ ( in which $L_{t+1}\in [t-L+2, t+1]$) for the next sliding window, shifted by one frame. 
Within the motion estimation, trajectory embeddings are applied to each layer. The output from the previous HyperSSM layer input to next layer, enabling precise refinement of the trajectory over time. Additionally, the hidden state acts as a messenger, transmitting information between cascaded layers, allowing the trajectory to progressively regress toward the ground truth labels.

\subsection{Overall Tracking Framework}
\label{overall}

\vspace{5pt} \noindent \textbf{Tracking Pipeline}
Our MOT approach adopts the tracking-by-detection paradigm, the overall tracking pipeline operates in five sequential stages: 

\textbf{(1) Detection:} a publicly YOLOX detector~\cite{YOLOX} processes each incoming frame to obtain candidate detections; 

\textbf{(2) Motion Estimation:} our \textit{HyperSSM} motion estimation module predicts the future positions of existing tracks; 

\textbf{(3) Tracks Association:} an adaptive association strategy that matches existing tracks with the most suitable detections from multiple confidence levels.

\textbf{(4) Tracks Initialization or Termination:} a track initialization strategy that selectively creates new tracks from unmatched high-confidence detections, tracks that remain unmatched for a predefined number of frames are either deleted or temporarily suspended.


\vspace{5pt} \noindent \textbf{Tracks Association.}
Instead of employing the conventional Hungarian algorithm for tracks association, we adopt a multi-stage adaptive association strategy as in ~\cite{tracktrack} that emphasizes track-level consistency. 
\textit{\textbf{Three categories of detections}} are constructed—high-confidence retained after Non-Maximum Suppression (NMS), low-confidence to recover missed targets, and high-confidence suppressed by NMS. For each track–detection pair, a cost based on motion, spatial, and appearance cues is computed.
The \textit{\textbf{matching process}} iteratively selects track–detection pairs with minimal mutual costs under adaptive thresholds, updating associations in a stage-wise manner until no valid matches remain.

\vspace{5pt} \noindent \textbf{Tracks Initialization or Termination.}
After association, unmatched high-confidence detections are considered for track initialization. 
To avoid redundant or false trajectories, we adopt a spatially aware initialization process: The latest positions of successfully matched tracks are treated as anchors, and both anchors and remaining detections undergo NMS to eliminate overlaps. 
Only detections that remain after suppression are used to create new tracks, ensuring that initialization occurs selectively and consistently. 

\begin{table}[t]\small
  \centering
  \setlength{\tabcolsep}{2mm}{
    \begin{tabular}{@{}lccccc@{}}
    \toprule
    Methods & HOTA & IDF1 & MOTA & AssA & DetA \\
    \midrule[0.5pt]
    \multicolumn{6}{l}{\textit{query-based methods:}}\\
    TransTrack~\cite{TransTrack}& 48.9 & 59.4 & 65.0 & 45.2 & 53.3 \\
    TransCenter~\cite{TransCenter} & 54.5 & 62.2& 73.2 & 49.7 & 60.1\\
    MOTR~\cite{MOTR}& 57.2 & 68.4 & 71.9 & 55.8 & /   \\
    MeMOT~\cite{MeMOT}& 56.9 & 69.0 & 72.5 & 55.2 & / \\
    MeMOTR~\cite{MeMOTR}& 58.8 & 71.5 & 72.8 & 58.4 & 59.6  \\
    MOTRv2~\cite{MOTRv2}&62.0&75.0&78.6&60.6&63.8\\
    MOTRv3~\cite{motrv3}&60.2&72.4&75.9&58.7&62.1\\
    MOTIP~\cite{motip}&59.3&71.3&75.3&57.0&62.0\\
    SAMBA~\cite{samba}&58.8&71.0&72.9&58.2&59.7\\
    DualMOT~\cite{dualMOT}&61.5&75.1&73.8&62.5&65.0\\
    \midrule[0.5pt]
    \multicolumn{6}{l}{\textit{motion-based methods:}}\\
    Bytetrack~\cite{ByteTrack}*& 63.1 & 77.3 & 80.3 & 62.0 & 64.5  \\
    StrongSORT~\cite{StrongSORT}*& 64.4 & 79.5 & 79.6 & 64.4 & 64.6 \\
    BoT-SORT~\cite{BoTSORT}*& 65.0 & 80.2 & 80.5 & 65.5 & 64.9 \\
    OC-SORT~\cite{OCSORT}*& 63.2 & 77.5 & 78.0 & 63.4 & 63.2 \\
    SparseTrack~\cite{sparsetrack}*& 65.1 & 80.1 & 81.0 & / & /\\
    HybridSORT~\cite{hybridSORT}*& 64.0 & 78.7 & 79.9 & / & /\\
    MambaMOT~\cite{mambatrackpp}&61.1&73.9&78.1&/&/ \\
    TrackSSM~\cite{trackssm}&61.4&74.1&78.5&59.6&63.6\\
    ETTrack~\cite{ettrack}&61.9&75.9&79.0&60.5&/\\
    DiffusionTrack~\cite{diffusiontrack}& 60.8 & 73.8& 77.9 & 58.8 & 63.2  \\
    OFTrack~\cite{OFTrack}&64.1&78.8&80.1&63.3&64.6 \\
    DiffMOT~\cite{diffmot}& 64.5 & 79.3 & 79.8 & 64.6 & 64.7 \\
    \textbf{HyperSSM (Ours)}& \textbf{66.9} & \textbf{83.1} & \textbf{81.5} & \textbf{68.2} & \textbf{66.3} \\
    \bottomrule
    \end{tabular}
    \caption{Performance comparison on \textbf{MOT17} test sets, following the private protocol. The \textbf{*} denotes Kalman Filter, others adopt learnable motion estimation.}
    \label{tab:mot17}
  }
\end{table}

\begin{table}[t]\small
  \centering
  \setlength{\tabcolsep}{2mm}{
    \begin{tabular}{@{}lccccc@{}}
    \toprule
     Methods& HOTA & IDF1 & MOTA & AssA & DetA\\
    \midrule[0.5pt]
    \multicolumn{6}{l}{\textit{query-based methods:}}\\
    TransCenter~\cite{TransCenter} & / & 58.7 & 67.7 & / & / \\
    MeMOT~\cite{MeMOT}& 54.1 & 66.1 & 63.7 & 55.0 & /  \\
    \midrule[0.5pt]
    \multicolumn{6}{l}{\textit{motion-based methods:}}\\
    Bytetrack~\cite{ByteTrack}*& 61.3 & 75.2 & 77.8 & 59.6 & 63.4  \\
    StrongSORT~\cite{StrongSORT}*& 62.6 & 77.0 & 73.8 & 64.0 & 61.3 \\
    BoT-SORT~\cite{BoTSORT}*& 63.3 & \textbf{77.5} & 77.8 & 62.9 & 64.0 \\
    OC-SORT~\cite{OCSORT}*& 62.4 & 76.3 & 75.7 & 62.5 & 62.4  \\
    SparseTrack~\cite{sparsetrack}*& 63.4 & 77.3 & 78.2 & / & /  \\
    HybridSORT~\cite{hybridSORT}*& 63.9 & 78.4 & 76.7 & / & /  \\
    DiffusionTrack~\cite{diffusiontrack}& 55.3 & 66.3 & 72.8 & 51.3 & 59.9  \\
    OFTrack~\cite{OFTrack}&63.4& 76.9 & 75.6 & 62.1 & 62.2  \\
    DiffMOT~\cite{diffmot}&61.7& 74.9 & 76.7 & 60.5 & 63.2  \\
    \textbf{HyperSSM (Ours)}&\textbf{65.2} & \textbf{77.5} & \textbf{80.0}& \textbf{66.3} & \textbf{62.5} \\
    \bottomrule
    \end{tabular}
    \caption{Performance comparison on \textbf{MOT20} test sets, following the private protocol. The \textbf{*} denotes Kalman Filter, others adopt learnable motion estimation.}
    \label{tab:mot20}
  }
\end{table}

\section{Experiments}

\subsection{Settings}

\textbf{Training loss.}
We take the smooth L1 loss and generalized-IoU~\cite{giou} between the estimation and ground-truth of all frames within the estimated sliding window:
\begin{equation}
 \mathcal{L} = \lambda_1 \sum_{t}^L \mathcal{L}_{L1}(P^t,P^t_{gt}) + \lambda_2 \sum_{t}^L \mathcal{L}_{GIoU}(P^t,P^t_{gt})
  \label{eq:loss}
\end{equation}
where $L$ represents the length of a sliding window, $\lambda_1$ and $\lambda_2$ represent the weight coefficients for smooth L1 loss and GIoU loss, respectively, $P_{gt}$ is the ground-truth labels.

\noindent\textbf{Dataset.}
We evaluate the performance of our HyperSSM in both linear motion and non-linear motion scenarios, linear motion corresponding to the MOT17~\cite{MOT16} and MOT20~\cite{MOT20} datasets, non-linear motion corresponding to the DanceTrack~\cite{DanceTrack} and SportsMOT~\cite{sportsmot} datasets. As most other trackers~\cite{ByteTrack, trackssm}, we merge the MOT17 and MOT20 training sets to train our model and report results on the MOT17 test set and MOT20 test set. For reporting results on the DanceTrack and SportsMOT test sets, we train HyperSSM separately on their respective training and validation sets. We conduct ablation experiments both on MOT17 and DanceTrack datasets, for MOT17 ablation, we train our model on the train-half set and ablation study test on the val-half set. For DanceTrack ablation, we train our model on the training sets and perform ablation testing on the validation set.

\noindent\textbf{Implementation Details.}
During the training phase of HyperSSM, trajectory segments—rather than images—are used for training, consistent with methodologies from DiffMOT~\cite{diffmot} and TrackSSM~\cite{trackssm}. The input consists of position and motion data derived from 5-frame trajectory segments. For training, we employ a batch size of 128, the Adam optimizer (learning rate \(= 0.0001\)), and an 8-layer flow decoder. Training is conducted over 360 epochs for the MOT17, MOT20, and SportsMOT datasets, and 260 epochs for DanceTrack. While the smooth L1 loss is used exclusively for MOT17, MOT20, and DanceTrack, SportsMOT training incorporates both smooth L1 and GIoU losses. This adjustment is driven by the larger inter-frame object displacement distances in SportsMOT, where GIoU loss enhances motion model convergence speed. 
All experiments are performed on NVIDIA Tesla V100 GPUs, implemented in PyTorch 1.22 and Python 3.11.


\begin{table}[t]\small
  \centering
  \setlength{\tabcolsep}{2mm}{
    \begin{tabular}{@{}lccccc@{}}
    \toprule
     Methods&HOTA & IDF1 & MOTA & AssA & DetA\\
    \midrule[0.5pt]
    \multicolumn{6}{l}{\textit{query-based methods:}}\\
    TransTrack~\cite{TransTrack}&45.5&45.2&88.4&27.5&75.9 \\
    MOTR~\cite{MOTR} &54.2&51.5&79.7&40.2&73.5  \\
    MeMOTR~\cite{MeMOTR}&68.5&71.2&89.9&58.4&80.5\\
    MOTRv2~\cite{MOTRv2}&73.4&76.0&92.1&64.4&83.7 \\
    MOTRv3~\cite{motrv3}&70.4&72.3&\textbf{92.9}&59.3&\textbf{83.8} \\
    MOTIP~\cite{motip}&72.0&76.8&91.9&63.5&81.8 \\
    SAM2MOT~\cite{sam2mot}&75.8&\textbf83.9&88.5&72.2& 79.7 \\
    SAMBA~\cite{samba}&67.2&70.5&88.1&57.5&78.8 \\
    DualMOT~\cite{dualMOT}&76.2&79.9&85.0&68.3&85.0 \\
    \midrule[0.5pt]
    \multicolumn{6}{l}{\textit{motion-based methods:}}\\
    StrongSORT~\cite{StrongSORT}*&55.6&55.2&91.1&38.6&80.7 \\
    BoT-SORT~\cite{BoTSORT}*&54.7&56.0&91.3&37.8&79.6\\
    Bytetrack~\cite{ByteTrack}*& 47.7&53.9&89.6&32.1&71.0\\
    OC-SORT~\cite{OCSORT}*&55.1&54.6&92.0&38.3&80.3\\
    SparseTrack~\cite{sparsetrack}*&55.7&58.1&91.3&39.3&79.2\\
    HybridSORT~\cite{hybridSORT}*&\textbf{65.7}&67.4&91.8&/&/\\
    DiffusionTrack~\cite{diffusiontrack} & 52.4 & 47.5 & 89.3 & 33.5 & 82.2 \\
    MambaTrack~\cite{mambatrack}&56.8&57.8&90.1&39.8&80.1 \\
    MambaMOT~\cite{mambatrackpp}&56.1&54.9&90.3&39.0&80.8\\
    TrackSSM~\cite{trackssm}&57.7&57.5&92.2&41.0&81.5\\
    ETTrack~\cite{ettrack}&56.4&57.5&92.2&39.1&81.7\\
    DiffMOT~\cite{diffmot}&62.3&63.0&\textbf{92.8}&47.2&\textbf{82.5}\\
    \textbf{HyperSSM(Ours)}&\textbf{65.7}&\textbf{68.0}&92.7& \textbf{52.6}&82.4\\
    \bottomrule
    \end{tabular}
    \caption{Performance comparison on \textbf{DanceTrack} test sets. The \textbf{*} denotes Kalman Filter, others adopt learnable motion estimation.}
    \label{tab:dancetrack}
  }
\end{table}

\begin{table}[t]\small
  \centering
  \setlength{\tabcolsep}{2mm}{
    \begin{tabular}{@{}lccccc@{}}
    \toprule
     Methods&HOTA & IDF1 & MOTA & AssA & DetA\\
    \midrule[0.5pt]
    \multicolumn{6}{l}{\textit{query-based methods:}}\\
    TransTrack~\cite{TransTrack}&68.9&72.5&92.6&57.5&82.7 \\
    MeMOTR~\cite{MeMOTR}&70.0&71.4&91.5&59.1&83.1\\
    MOTIP~\cite{motip}&72.6&77.1&92.4&63.2&83.5 \\
    SAMBA~\cite{samba}&69.8&71.9&90.3&59.4&82.2 \\
    DualMOT~\cite{dualMOT}&73.9&78.7&91.5&66.6&82.2 \\
    \midrule[0.5pt]
    \multicolumn{6}{l}{\textit{motion-based methods:}}\\
    Bytetrack~\cite{ByteTrack}*&64.1&71.4&95.9&52.3&78.5\\
    OC-SORT~\cite{OCSORT}*&73.7&74.0&96.5 &61.5&88.5\\
    MambaTrack~\cite{mambatrack}&72.6&72.8&95.3&60.3&87.6 \\
    MambaMOT~\cite{mambatrackpp}&71.3&71.1&94.9&58.6&86.7 \\
    TrackSSM~\cite{trackssm}&74.4&74.5&96.8&62.4&88.8\\
    ETTrack~\cite{ettrack}&74.3&74.5&96.8&62.1&88.8\\
    MixSort~\cite{sportsmot}&74.1&74.4&96.5&62.0&88.5\\
    DiffMOT~\cite{diffmot}&76.2&76.1&97.1&65.1&89.3\\
    \textbf{HyperSSM(Ours)}&\textbf{78.5}&\textbf{80.0}&\textbf{97.0}&\textbf{69.2}&\textbf{89.2}\\
    \bottomrule
    \end{tabular}
    \caption{Performance comparison on \textbf{SportsMOT} test sets. The \textbf{*} denotes Kalman Filter, others adopt learnable motion estimation.}
    \label{tab:sportsmot}
  }
\end{table}

\subsection{SOTA Comparison}

\noindent\textbf{MOT17} and \textbf{MOT20} are datasets for multi-pedestrian tracking under the linear motion pattern. We use the standard split and obtain the test set evaluation by submitting the results to the online MOTChallenge website. 
The performances are presented in Table \ref{tab:mot17} and \ref{tab:mot20} under the "private" protocol. Our HyperSSM achieves a superior performance both in MOT17 and MOT20 with the HOTA of 66.9 and 65.2, showing impressive performance among all models. 

\noindent\textbf{DanceTrack} is a long-range group dancing dataset and has similar disruptors, severe occlusion, and frequent crossovers with highly complex non-linear motion. Table~\ref{tab:dancetrack} shows that our HyperSSM performs superior quality, obtains the and 65.7 HOTA and 68.0 IDF1 score.

\noindent\textbf{SportsMOT} represents sports environments and is characterized by varied motion patterns and intricate non-linear movements. As illustrated in Table~\ref{tab:sportsmot}, our HyperSSM outperforms other trackers with a notable HOTA score of 78.5, highlighting its superior performance.

\subsection{Ablation Study}

We ablate our approach using the MOT17 and DanceTrack datasets. For MOT17, we follow the methodology of ByteTrack by splitting the MOT17 training set into a train-half set and a val-half set. For DanceTrack, we are trained on full training set and tested on validation set.

\begin{table}[t]\small
  \centering
  \begin{tabular}{@{}c@{ }|c@{\quad}c@{\quad}c|c@{\quad}c@{\quad}c@{}}
    \toprule
    \multirow{2}*{\makecell{Motion\\methods}} & \multicolumn{3}{c}{MOT17} & \multicolumn{3}{c}{DanceTrack}\\
     &HOTA &IDF1 & MOTA &HOTA &IDF1 & MOTA\\
    \midrule[0.5pt]
    Kalman& 64.5& 74.7& 79.1& 47.0 &52.2 &87.3\\
    OC-Kalman& 64.6& 74.9& 79.0& 52.5&54.3 & 89.1\\
    SSM& 63.4& 72.1& 78.6&53.7 &53.5 & 88.9\\
    \rowcolor{gray!20}
    HyperSSM & 65.5 & 75.9 & 79.7 & 59.7 & 60.6 & 89.5\\
    \bottomrule
    \end{tabular}
  \captionof{table}{Comparison with other motion models. The detection and association strategies are fixed across all methods.}
  \label{tab:ablation1}
\end{table}

\noindent\textbf{Comparison with other motion models.} By maintaining consistent detection and hyper-parameter settings, we replace our estimation module HyperSSM with Kalman Filter, OC-Kalman~\cite{OCSORT}, and traditional SSM~\cite{trackssm} to assess performance differences. Results in Table~\ref{tab:ablation1} show that the Kalman series excels in MOT17 due to its constant velocity assumption. HyperSSM offers comparable gains over the Kalman filter in MOT17, but delivers substantially larger improvements in non-linear patterns. Additionally, HyperSSM consistently surpasses traditional SSM across all scenarios, demonstrating broader applicability.

\begin{figure}
  \centering
  \begin{subfigure}{0.48\linewidth}
    \includegraphics[width=1\linewidth,keepaspectratio,page=5]{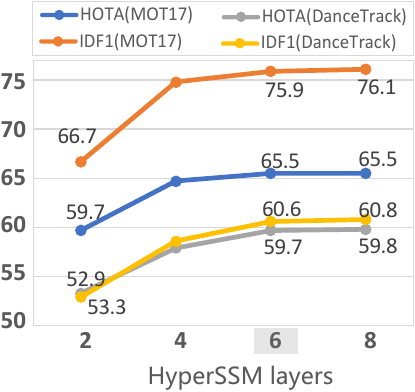}
    \vspace{-10pt}
  \end{subfigure}
  \begin{subfigure}{0.48\linewidth}
    \includegraphics[width=1\linewidth,keepaspectratio,page=4]{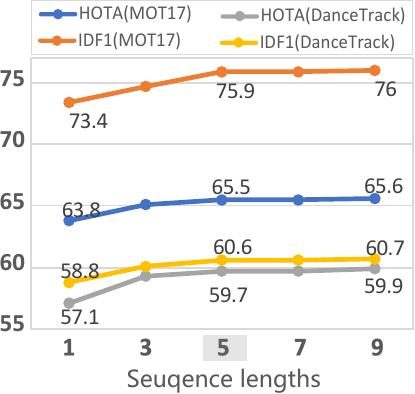}
    \vspace{-10pt}
  \end{subfigure}
  \vspace{-10pt}
  \caption{Ablation study on the number of HyperSSM layers(left) and the length of trajectory segments(right).}
  \label{fig:aba2}
\end{figure}

\noindent\textbf{HyperSSM layers and sequence length in motion estimation.}
In our motion estimation module, we analyze two key design factors: the number of HyperSSM layers and the length of trajectory segments. As shown in the left part of~\ref{fig:aba2}, increasing the number of layers yields diminishing returns, with improvements becoming negligible beyond 8 layers; thus, we select 8 iterations as a balanced choice between accuracy and efficiency. We further examine the influence of historical segment length by evaluating trajectories of 1, 3, 5, 7, and 9 frames. Although longer segments improve predictive capability, excessively long histories can extend beyond the optimal prediction horizon. As shown in the right part of ~\ref{fig:aba2}, performance peaks at a segment length of 5 frames and we adopt this configuration in our final framework.




\begin{table}[t]
  \centering
  \begin{tabular}{c|cc>{\columncolor{gray!30}}cccc}
    \toprule
     $\theta$& 0.5 & 0.7 & 0.8 & 0.9 & 0.95 & 0.98 \\
    \midrule[0.5pt]
    HOTA& 47.7 & 62.1 & 65.5 & 65.3 & 64.8 & 64.7\\
    IDF1& 47.3 & 68.8 & 75.9 & 75.5 & 73.4 & 72.9\\
    MOTA& 72.1 & 76.2 & 79.7 & 79.5 & 79.4 & 77.8\\
    \bottomrule
    \end{tabular}
    \captionof{table}{Ablation analysis of parameter $\theta$, it measures motion similarity between two objects to establish hyperedges.}
    \label{tab:ablation4}
\end{table}

\noindent\textbf{Ablation analysis of critical hyperparameter.}
The parameter $\theta$ measures the motion similarity between two objects and determines whether a hyperedge is established between them. 
We evaluate $\theta$ on MOT17 in Table~\ref{tab:ablation4}, the best performance is achieved when $\theta = 0.8$ (HOTA = 65.5). 
When the threshold increases beyond 0.8, performance drops because overly strict constraints reduce the number of valid associations. 
This observation is intuitive: a smaller threshold introduces excessive associations, while a larger threshold results in too few connections.

\begin{table}[t]
  \centering
  \begin{tabular}{l|c|cccc}
    \toprule
    Model & SSM & \multicolumn{4}{c}{HyperSSM}\\
    YOLOX & X & X & L & M & S \\
    \midrule[0.5pt]
    HOTA& 71.9 & 74.1 & 72.4 & 68.1 & 57.3\\
    IDF1& 77.0 & 81.9 & 78.8 & 73.7 & 67.0\\
    \midrule[0.1pt]
    Params(M)& 104.3 & 135.3 & 76.5 & 41.1 & 19.0 \\
    \textit{fps}& 20.3 & 17.5 & 22.7 & 26.8 & 28.8 \\
    \bottomrule
    \end{tabular}
    \captionof{table}{Efficiency comparison on various YOLOX models.}
    \label{tab:ablation4}
\end{table}

\noindent\textbf{Efficiency comparison.}
Table \ref{tab:ablation4} presents an efficiency comparison among various YOLOX models and the original SSM model. Notably, the pretrained YOLOX models utilized in this comparison are trained on the complete MOT17~\cite{MOT16} train set. To mitigate the impact of pre-training data, both training and inference operations are conducted on the MOT17 full train set rather than the half train set. It is essential to recognize that the experimental results in this context serve as a reference for comparing model sizes and do not adhere to the same evaluation standards as other results. The efficiency findings reveal that the lightweight model achieves a higher tracking speed, whereas the larger model demonstrates superior performance.



\subsection{Qualitative analysis on motion challenges}
\begin{figure}
  \centering
  \begin{subfigure}{1\linewidth}
    \includegraphics[width=1\linewidth,keepaspectratio,page=6]{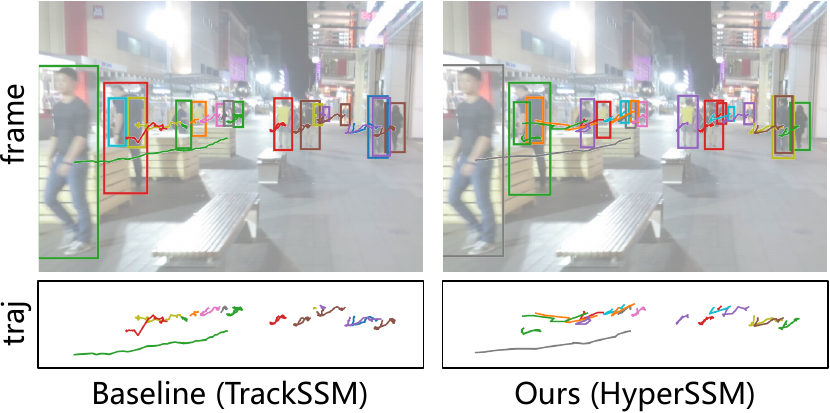}
  \end{subfigure}
  \begin{subfigure}{1\linewidth}
    \includegraphics[width=1\linewidth,keepaspectratio,page=7]{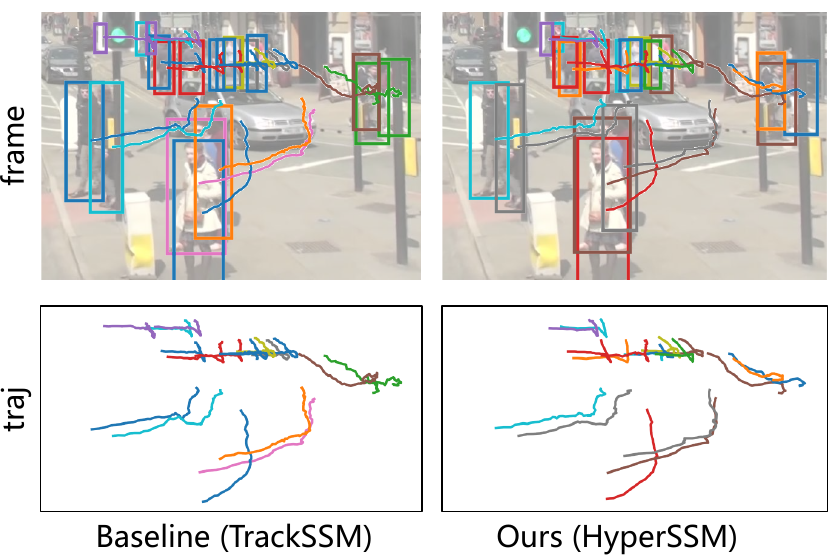}
  \end{subfigure}
  \vspace{-10pt}
  \caption{Visualizations under the two key challenge scenarios: a motion-smoothness–dependent scene (top) and a crowded scene with severe occlusion (bottom).}
  \label{fig:vis}
\end{figure}

To further demonstrate the effectiveness of our method, we visualize motion estimation trajectories on representative MOT17 scenes (Figure~\ref{fig:vis}) that correspond to the two key challenges.
For \textbf{motion instability (C1)}, we use a surveillance sequence with linear pedestrian motion. The baseline produces trajectories with clearly observable fluctuations, indicating its sensitivity to noise. In contrast, our method generates much smoother and more stable trajectories. This result shows that collaborative reasoning, achieved by jointly estimating motion across multiple correlated targets, can effectively reduce noise in individual motion predictions.
For \textbf{occlusion robustness (C2)}, we visualize a crowded street scenario with frequent and severe occlusions. Compared with the baseline, our method exhibits fewer trajectory drifts and preserves identity continuity more accurately. These observations confirm that the collaborative design provides stronger robustness in heavily occluded environments.


\section{Conclusion}
In this work, we introduced HyperSSM, a collaborative motion estimation framework that leverages hypergraph-based interaction modeling together with a state space formulation. Through the collaborative reasoning mechanism enabled by HyperSSM, multiple objects can inform and refine one another’s motion estimates, which effectively suppresses noise fluctuations and reduces the impact of occlusion. The integration of hypergraph relational constraints with a learnable state propagation process allows the model to capture structured spatial relations while maintaining stable temporal dynamics.
Beyond achieving strong tracking accuracy, our study reveals that incorporating high-order relationships among objects is a powerful principle for motion estimation, offering a new direction that moves beyond independent per-object prediction. The experimental results across several mainstream benchmarks confirm the reliability and robustness of this design.
Future work may extend HyperSSM to broader tracking scenarios or integrate richer modalities to further enhance collaborative motion reasoning.

%% file: sec/3_finalcopy.tex
\section*{Acknowledgement}

This work is supported by the National Natural Science Foundation of China (Numbers 62272184 and 62402189), 
the China Postdoctoral Science Foundation (Numbers 2024M751012, 2025T180429, and GZC20230894),  
the Postdoctor Project of Hubei Province (Number 2024HBBHCXB014),
and the National Research Foundation, Singapore, under its Medium Sized Center for Advanced Robotics Technology Innovation.